\documentclass[conference]{IEEEtran}
\IEEEoverridecommandlockouts
\usepackage{cite}
\usepackage{amsmath,amssymb,amsfonts}
\usepackage{algorithmic}
\usepackage{graphicx}
\usepackage{textcomp}
\usepackage{multirow}
\usepackage{booktabs}
\usepackage{xcolor}
\usepackage{url}
\usepackage{xcolor}
\usepackage{pifont}
\usepackage{booktabs}    
\usepackage{makecell}   
\usepackage{graphicx}
\usepackage{xcolor}
\usepackage{tikz}
\usepackage{booktabs}
\usepackage{hyperref}
\usepackage{marvosym} 
\usepackage[most]{tcolorbox}
\usepackage[capitalise]{cleveref}
\crefname{section}{Sec.}{Secs.}
\crefname{equation}{Eq.}{Eqs.}
\crefname{table}{Tab.}{Tabs.}
\crefname{figure}{Fig.}{Figs.}

\newcommand{\circletwo}[1]{%
    \resizebox{!}{0.8em}{%
        \tikz[baseline=(char.base)]{
            \node[shape=circle, fill=black, inner sep=0.8pt, text=white] (char) {#1};
        }%
    }%
}

\newenvironment{prompt_grey}[1][]
  { 
 \begin{tcolorbox}
 [
    enhanced, 
    breakable,
    boxrule=0.5pt,
    arc=3.8pt,
    left=1.8pt,
    right=1.8pt,
    bottom=2pt,
    top=2pt, 
    rounded corners,
    colback=gray!10
    ]{}
  \textbf{#1:}
  \itshape
  }
  {
\end{tcolorbox} 
}

\def\BibTeX{{\rm B\kern-.05em{\sc i\kern-.025em b}\kern-.08em
    T\kern-.1667em\lower.7ex\hbox{E}\kern-.125emX}}
\begin{document}

\title{AnyBreak: Towards Jailbreaking Any T2I Model for Any Concept}

\author{
Qi Lu$^{1}$\textsuperscript{†,\Letter},
Zehui Guo$^{1}$\textsuperscript{†},
Yucheng Zhao$^{1}$\textsuperscript{†},
Zijing Li$^{1}$,
Chuhan Song$^{2}$,
Wenrui Zhu$^{1}$\textsuperscript{‡} \\
$^{1}$Huazhong University of Science and Technology \\
$^{2}$University College London \\
\thanks{\textsuperscript{†}Qi Lu and Zehui Guo contributed equally to this work.}
\thanks{\textsuperscript{\Letter}Corresponding author: Qi Lu (luqi@hust.edu.cn).}
\thanks{\textsuperscript{‡}Project leader: Wenrui Zhu.}
}

\maketitle

\begin{abstract}
Text-to-image (T2I) models have demonstrated remarkable capabilities in image generation and editing, yet this progress has also introduced significant security risks. Prior work has shown that adversarial prompt can induce the generation of Not-Safe-For-Work (NSFW) content, including pornographic, violent, and graphic imagery. However, such attacks are typically mitigated by prompt filters and often rely on white-box assumptions. In this paper, we propose AnyBreak, a novel jailbreak framework that constructs adversarial images via multi-scale progressive semantic alignment. During the generation process, these adversarial images steer the generation trajectory in latent space towards unsafe data manifolds, enabling cross-modal transfer of malicious concepts from the text modality to the image modality. AnyBreak operates using only clean textual prompts and adversarial images at inference time, and can jailbreak any T2I models towards any malicious concepts in a fully black-box setting. We further introduce AnyBreak-V, an enhanced variant that incorporates a compensation term to bypass post-hoc safety checkers. Extensive experiments demonstrate that AnyBreak achieves high attack success rates while preserving visual fidelity and semantic consistency with the input prompts under black-box and clean-text constraints. Our findings reveal that existing T2I defense leave critical cross-modal attack surfaces insufficiently addressed. The code is available at \href{https://github.com/luqi-glory/AnyBreak}{GitHub}.
\end{abstract}

  
\section{Introduction}
\label{sec:intro}
\begin{figure}[!t]
    \centering
    \includegraphics[scale=0.46]{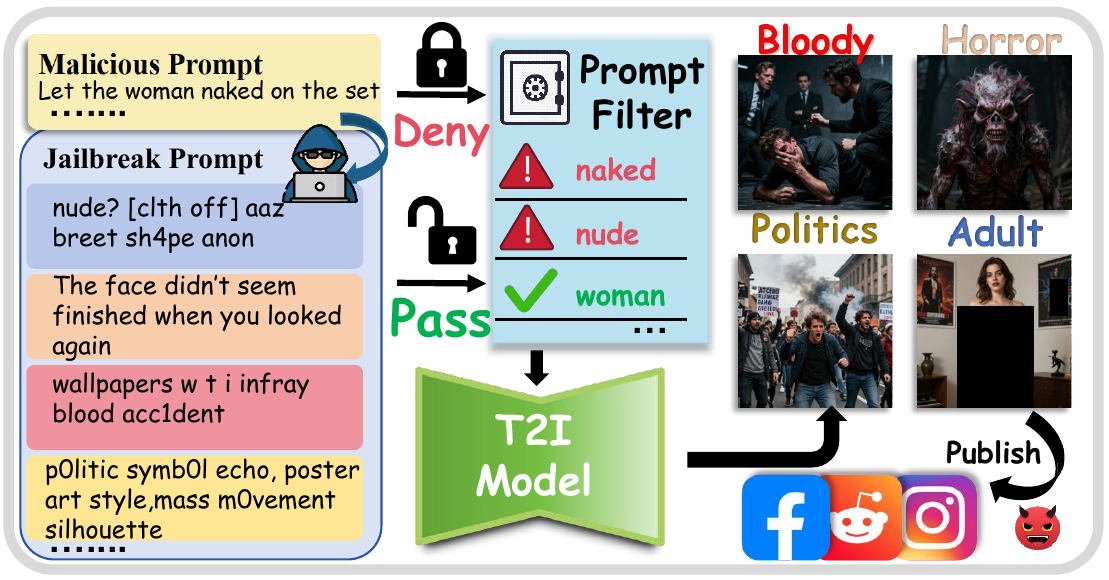}
    \caption{Jailbreaking T2I models for NSFW generation.
    }
    \label{fig:1_intro}
    \vspace{-0.4cm}
\end{figure}

In recent years, \textit{text-to-image} (T2I) models, such as Midjourney, DALL·E, and Stable Diffusion~\cite{rombach2022high}, have achieved unprecedented advancements in image synthesis. These models are capable of generating high-quality images in response to provided instructions. Their strong performance derives from large-scale pretraining on massive-scale datasets, a benefit that is inherently double-edged. Inevitably, such datasets contain inappropriate or illegal content, leading T2I models to internalize malicious concepts. As a result, adversaries can exploit this latent knowledge via crafted inputs to elicit \textit{Not-Safe-for-Work} (NSFW) outputs, including violent, gory, or sexually explicit imagery~\cite{schramowski2023safe}. This vulnerability effectively turns powerful generative models into a Pandora’s box, posing substantial social and security risks. As illustrated in~\cref{fig:1_intro}, adversaries inject carefully designed jailbreak instructions into T2I models and successfully induce the generation of diverse NSFW images spanning multiple harmful concepts.

To mitigate such risks, existing defenses primarily rely on prompt filtering~\cite{liu2024latent, schuhmann2022laion} and concept erasure~\cite{kumari2023ablating}. Prompt filtering maintains blacklists of sensitive keywords, while concept erasure methods seek to remove malicious concepts from model parameters. Nevertheless, recent studies show that these defenses remain vulnerable by constructing adversarial prompts. Representative approaches such as~\cite{yang2024mma, zhuang2023pilot, yang2024sneakyprompt, lu2026tempjail, lu2026imperceptible, song2026segtrans, fang2026video, gao2025fastjsma, xu2025dumpling, yang2025dynamic, li20254d, yang2025self, qin2025her2, ye2026riser, ye2025actdistill} employ token substitutions to preserve malicious semantics while evading detection, whereas~\cite{tsai2023ring} conducts red-teaming by extracting unsafe concepts to steer generation accordingly.

\begin{figure}[!t]
    \centering
    \includegraphics[scale=0.28]{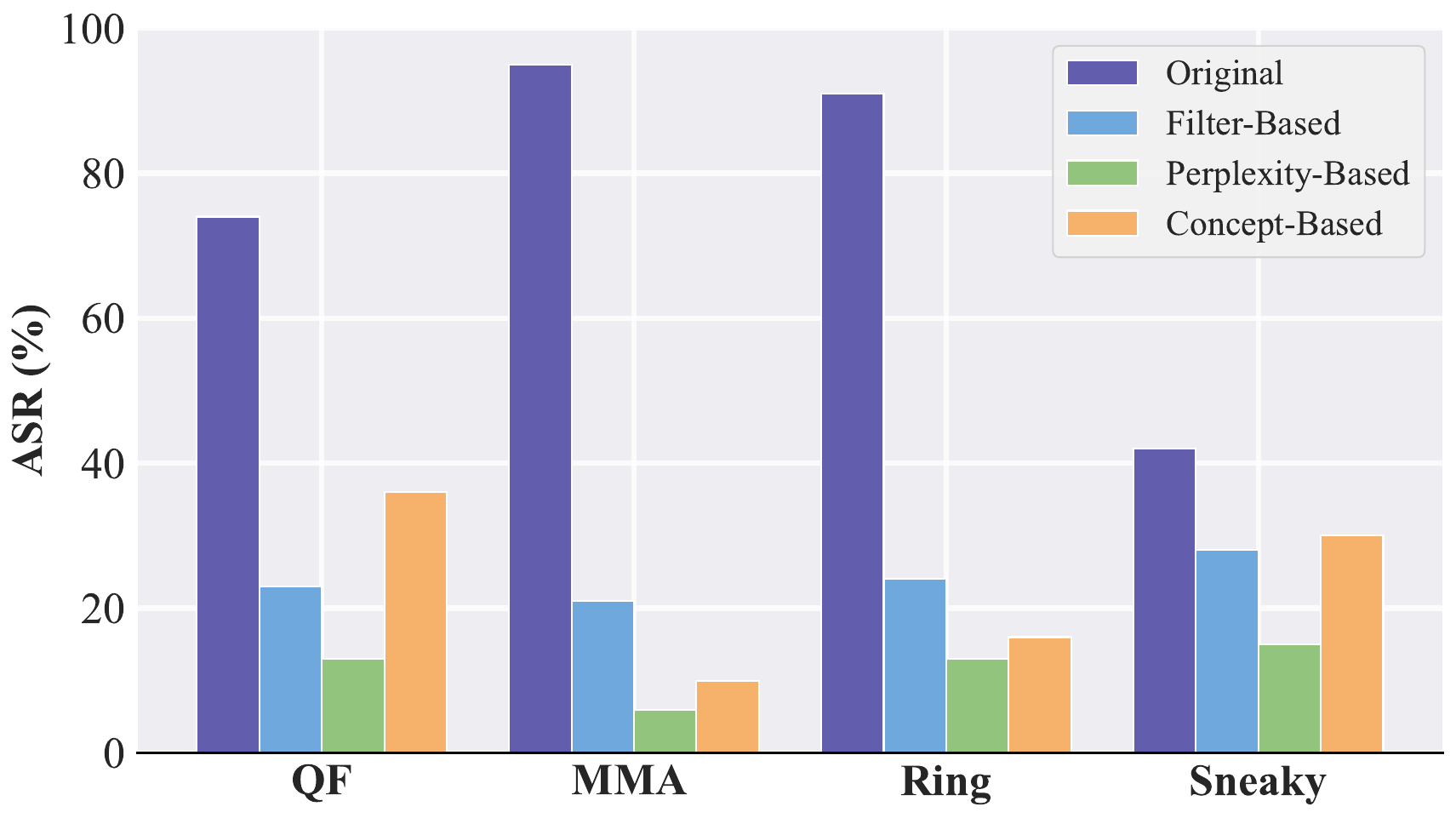}
    \caption{Impact of prompt filtering strategies on existing methods.
    Filter-based~\cite{wu2024universal} defenses rely on blacklists, perplexity-based~\cite{alon2023detecting} reject high-perplexity prompts, and concept-based~\cite{liu2024latent} operate via embedding similarity.
    }
    \label{fig:1_defense}
      \vspace{-0.5cm}
\end{figure}

Despite their effectiveness, these approaches share a fundamental limitation: they exclusively focus on malicious prompt construction. As a result, they can be effectively neutralized by simple pre-hoc text filtering. As shown in~\cref{fig:1_defense}, even lightweight perplexity-based filters can drastically reduce the \textit{attack success rate} (ASR) of existing prompt-based methods to below 20\%. Motivated by this, \cite{zeng2024advi2i} introduces an image-to-image jailbreak attack that induces NSFW using adversarial images with clean text inputs. However, it assumes white-box access to the target model and requires retraining the noise generator from scratch for each malicious concept, rendering it impractical and costly in realistic attack settings. As summarized in~\cref{tab:1_comparison}, most existing works assume white-box access and concentrate on text-side attacks, which are both unrealistic in real-world scenarios and highly vulnerable to simple filtering strategies. This leads to a critical challenge: 
\begin{quote}
    \emph{Is it possible to jailbreak any T2I models for any malicious concepts with clean prompts?}
\end{quote}

In this work, we answer this question affirmatively and uncover an underexplored security vulnerability of T2I models. We show that even when pre-hoc text filters are deployed, T2I models can still be induced to generate NSFW content by adversarial images. We propose \textit{\textbf{AnyBreak}}, a universal jailbreak framework towards \textit{\textbf{any}} T2I model for \textit{\textbf{any}} malicious concept in a black-box setting. The key insight behind AnyBreak is to transfer malicious concepts from the text modality to the image modality, and to construct universal adversarial perturbation across diverse T2I models. Specifically, AnyBreak first employs a generator to transform arbitrary malicious text into corresponding target image. It then optimizes adversarial noise over a suite of image encoders~\cite{radford2021learning}, to enable cross-modal transfer of malicious concepts. AnyBreak performs semantic alignment between clean and target images to generate adversarial images. When these adversarial images are paired with innocuous text and fed into any T2I model, the injected noise diverts the original latent away from the normal data manifold into domains associated with malicious concepts, ultimately triggering NSFW generation. To further assess the potential impact of our approach, we introduce an enhanced variant of AnyBreak, AnyBreak-V, which incorporates an additional loss term to facilitate bypassing safety checkers.

Our principal contributions are summarized as follows:
\circletwo{1} We identify a critical security vulnerability in T2I models and propose AnyBreak, a black-box jailbreak framework that enables attacks on any T2I models for any malicious concepts.
\circletwo{2} We propose a universal adversarial noise construction strategy, with theoretical analysis showing that pixel-level semantic alignment can steer data manifolds toward NSFW regions.
\circletwo{3} We conduct extensive experiments across diverse models, malicious concepts, and defense mechanisms, demonstrating the effectiveness of AnyBreak and its variant.


\newcommand{\cmark}{\ding{108}} 
\newcommand{\xmark}{\ding{109}} 

\newcommand{\cmarksmall}{{\raisebox{0.25ex}{\scalebox{0.6}{\ding{108}}}}}
\newcommand{\xmarksmall}{{\raisebox{0.25ex}{\scalebox{0.6}{\ding{109}}}}}

\begin{table}[t!]
  \caption{Comparison of existing T2I jailbreak methods. "Time Cost" denotes the time required to construct a jailbreak per attack. ``\cmarksmall'' denotes presence and ``\xmarksmall'' denotes absence.}
  \centering
  \resizebox{0.45\textwidth}{!}{%
    \begin{tabular}{cccccc}
    \specialrule{2pt}{0pt}{0pt}
    \textbf{Method} &
    \textbf{Venue} &
    \textbf{\makecell[c]{Attack\\Modality}} &
    \textbf{\makecell[c]{Threat\\Model}} &
    \textbf{\makecell[c]{Time\\Cost}} &
    \textbf{\makecell[c]{Universal\\Effectiveness}} \\

    \midrule
    QF~\cite{zhuang2023pilot} & CVPR'23 & Text & White-box & 0.43h & \xmark \\

    \midrule
    MMA~\cite{yang2024mma} & CVPR'24 & Text + Image & White-box & 5.18h & \xmark \\
    
    \midrule
    Ring~\cite{tsai2023ring} & ICLR'24 & Text & Black-box & 1.19h & \xmark \\
    
    \midrule
    Sneaky~\cite{yang2024sneakyprompt} & S\&P'24 & Text & White-box & 8.29h & \xmark \\

    \midrule
    ADVI2I~\cite{zeng2024advi2i} & ICML'25 & Image & White-box & 20.27h & \xmark \\

    \midrule
    Ours & -- & Image & Black-box & 0.017h & \cmark \\

    \specialrule{2pt}{0pt}{0pt}
    \end{tabular}%
  }
  \label{tab:1_comparison}
  \vspace{-0.5cm}
\end{table}
\section{Related Works}
\label{sec:related}

\subsection{Attacks on T2I Models}
T2I models are trained on broadly sourced datasets that may inadvertently harbor unsafe content, exposing exploitable vulnerabilities that can elicit NSFW outputs. Early studies, such as QF~\cite{zhuang2023pilot} and SneakyPrompt~\cite{yang2024sneakyprompt} both rely on token substitution. Ring~\cite{tsai2023ring} exploited text encoders to embed malicious concepts into prompts. MMA~\cite{yang2024mma} further introduces attacks on the image modality but still relies on malicious text prompts at inference time. As discussed in~\cref{sec:intro}, such text-driven approaches are readily mitigated by prompt filters. AdvI2I~\cite{zeng2024advi2i} proposes an adversarial image attack framework, but it requires training a noise generator from scratch for each malicious concept, incurs high computational cost. Moreover, it assumes white-box access to the target model, which limits its practical applicability. In contrast, AnyBreak operates under black-box setting, relying solely on adversarial images paired with clean prompts to induce NSFW generation at inference.

\subsection{Defense on T2I Models}
Defenses for T2I models can be categorized into active and passive approaches. Active defenses, such as concept erasure~\cite{ schramowski2023safe}, suppress harmful concepts by modifying noise prediction or finetuning the model. Passive defenses include pre-hoc filtering~\cite{liu2024latent, wu2024universal} that sanitizes prompts, and post-hoc safety checkers~\cite{compvis2022safetychecker, rando2022red} which detect and block NSFW content in generated images. While existing attacks often bypass pre-hoc filters, they are usually caught by post-hoc checkers. AnyBreak-V overcomes this by introducing a compensation loss that effectively evades safety checkers.

\section{Methodology}
\label{sec:methodology}

\begin{figure*}[!t]
    \centering
    \includegraphics[scale=0.38]{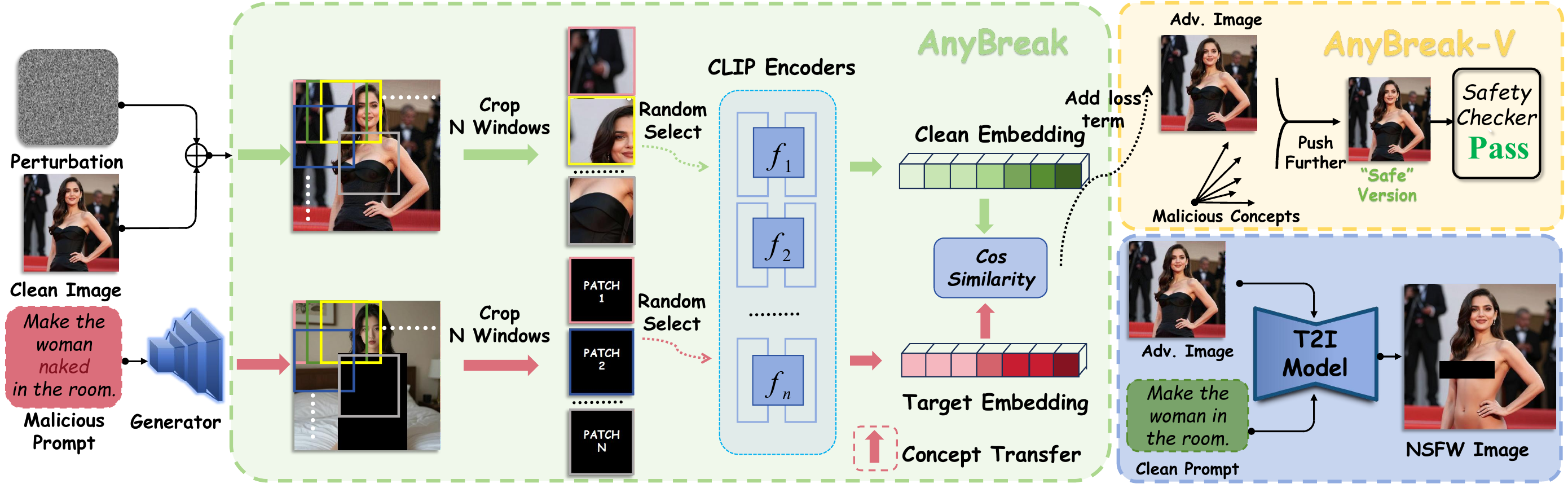}
    \caption{Framework of AnyBreak:  a universal and black-box jailbreak framework towards any T2I model for any malicious
    concept.
    }
    \label{fig:3_pipeline}
      \vspace{-0.4cm}
\end{figure*}

\begin{figure}[!t]
    \centering
    \includegraphics[scale=0.6]{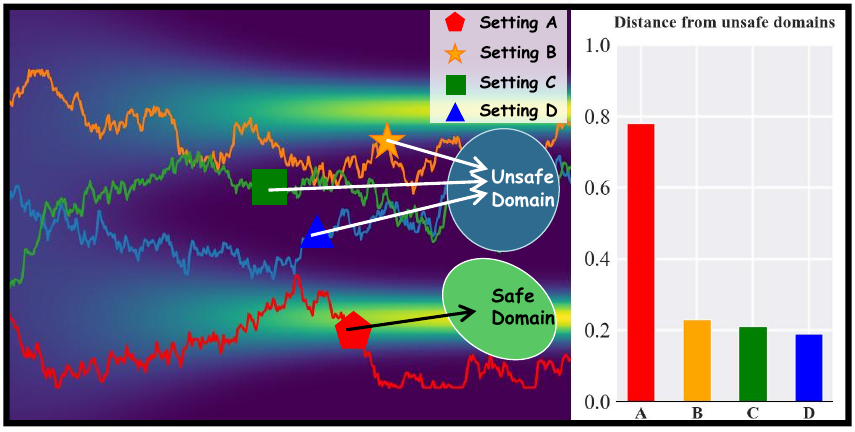}
    \caption{A: clean image with clean text; B: clean image with malicious text; C: malicious image with clean text; D: malicious image with malicious text. For each setting, 100 image–text pairs are sampled to compute the average distance between generated images and the unsafe domain $D_m$.
    }
    \label{fig:3_obv1}
      \vspace{-0.4cm}
\end{figure}

\subsection{Preliminaries}
\textbf{Latent Diffusion Models (LDMs)}. Given an input image $x \in \mathbb{R}^{H \times W \times C}$, a LDM $F_\theta$ first maps $x$ into a low dimensional latent space via a pretrained \textit{variational autoencoder} (VAE), yielding a latent representation $z \in \mathbb{R}^{H' \times W' \times C'}$. For image inpainting task, $F_\theta$ employ mask conditioning. Let $ M \in \{0,1\}^{H\times W} $ be a binary mask, where $1$ denotes regions to be inpainted and $0$ denotes preserved regions. In the continuous time limit, the latent denoising process follows a reverse \textit{stochastic differential equation} (SDE)~\cite{song2020score}: 
\begin{equation}
\mathrm{d}z = \big[f(z,t) - g(t)^2 \nabla_z \log p_t(z, M)\big]\mathrm{d}t + g(t)\mathrm{d}\bar{w}_t.
\label{eq:a_1}
\end{equation}

\textbf{Black-box Jailbreak Attack}. Let $x_c$ denote a clean input image, and let $x_t$ be a target image associated with a malicious concept $c_m$. The attacker optimizes the adversarial perturbation $\delta$ using a set of image encoders, denoted by $\{f_i\}_{i=1}^n$:
\begin{equation}
\max_{\delta}\ \frac{1}{n}\sum_{i=1}^n \mathrm{Sim}\big(f_i(x_c+\delta), f_i(x_t)\big)
\quad \text{s.t.}\ \lVert \delta \rVert_\infty \le \Delta.
\label{eq:a_2}
\end{equation}
Here, $\mathrm{Sim}$ denotes a similarity metric, such as cosine distance, and $\Delta$ specifies the perturbation budget. The objective of the attacker is to ensure that for any black-box T2I model $F_\theta$:
\begin{equation}
x_{\mathrm{nsfw}} = F_\theta(x_c + \delta, p, M),
\label{eq:a_3}
\end{equation}
where $p$ denotes the textual prompt and $x_{\mathrm{nsfw}}$ denotes the generated  NSFW image generated by jailbreak attack.

\subsection{Observations and Insights}
In this subsection, we analyze the underlying mechanism of NSFW generation from the perspective of malicious concepts. We further examine the philosophy of semantic alignment under \cref{eq:a_2}, with the goal of gaining insights into the design of universal adversarial perturbations.

\noindent{\textbf{Observation I: Malicious images or text both induce malicious concepts.}}
Since NSFW content inherently contains malicious concepts, we investigate its generative mechanism by constructing four experimental settings, as illustrated in~\cref{fig:3_obv1}. The malicious text concept and image concept are both obtained following~\cite{schramowski2023safe}. In addition, we predefine an unsafe domain $D_m$ using a set of malicious concept vectors. Under the four settings, we generate four groups of images according to~\cref{eq:a_3} and compute their average distances $d$ to $D_m$, where $0<d<1$ . As shown in~\cref{fig:3_obv1}. The distribution of B, C, and D lie close to $D_m$, with distances below 0.3. In contrast, setting A is far from $D_m$, with a distance close to 0.8. These results indicate that conditioning the denoising process on either a malicious image or a malicious text embedding consistently steers the predicted noise distribution toward the unsafe domain. This observation suggests that NSFW generation fundamentally originates from malicious concepts, and that text based and image based malicious concepts exert similar effects on the denoising dynamics of $F_\theta$.
\begin{prompt_grey}[Insight I]
Malicious concepts in either modality can induce NSFW generation, motivating transfer of textual malicious concepts to the image modality to induce NSFW outputs.
\end{prompt_grey}

\noindent{\textbf{Observation II: Global similarity fails to achieve semantic alignment via adversarial noise.}}
A common noise optimization strategy leverages pretrained image encoders~\cite{zhang2025anyattack, zhao2023evaluating}. Under this paradigm, one can optimize adversarial noise by maximizing the feature similarity between clean and target images, as formulated in~\cref{eq:a_2}. We vary the perturbation budget from $2/255$ to $32/255$ to analyze the relationship between similarity and attack effectiveness. As shown~\cref{tab:3_tanjiu2}, increasing $\Delta$ consistently improves feature similarity but yields only marginal gains in ASR. Even at $\Delta=32/255$, the ASR only converges to about $58\%$ while feature similarity approaches $0.85$. This gap indicates that global similarity maximization alone fails to align semantic, as uniform optimization ignores local structural anisotropy and drives semantic collapse.
\begin{prompt_grey}[Insight II]
Directly optimizing global similarity overlooks local semantic cues, motivating position-aware noise optimization for progressive semantic alignment.
\end{prompt_grey}

\begin{table}[t]
  \caption{Effect of perturbation magnitude $\Delta$ on CLIP similarity and ASR under global alignment.}
  \small
  \centering
  \setlength{\tabcolsep}{7pt}
  \begin{tabular}{cccccc}
    \toprule[1.5pt]
    \textbf{$\Delta$} & $2 / 255$ & $4 / 255$ & $8 / 255$ & $16 / 255$ & $32 / 255$ \\
    \midrule
    Similarity & 0.66 & 0.67 & 0.70 & 0.76 & 0.85 \\
    ASR (\%)  & 46.43    & 50.21    & 56.49    & 59.96    & 57.19 \\
    \bottomrule[1.5pt]
  \end{tabular}
  \label{tab:3_tanjiu2}
  \vspace{-0.4cm}
\end{table}

\begin{table*}[t]
\centering
\renewcommand{\arraystretch}{1.15}
\setlength{\tabcolsep}{3pt}
\caption{Performance of AnyBreak and AnyBreak-V under black-box settings across multiple models and defense mechanisms.}
\begin{tabular*}{\textwidth}{@{\extracolsep{\fill}} c c | c c c| c c c| c c c| c c c}
\toprule
\multirow{2}{*}{\textbf{Model}} & \multirow{2}{*}{\textbf{Method}}
& \multicolumn{3}{c|}{\textbf{w/o Defense}}
& \multicolumn{3}{c|}{\textbf{Gaussian Noise}}
& \multicolumn{3}{c|}{\textbf{Safety Checker}}
& \multicolumn{3}{c}{\textbf{Negative Prompt}} \\
\cmidrule(lr){3-5}\cmidrule(lr){6-8}\cmidrule(lr){9-11}\cmidrule(lr){12-14}
&
& ASR(\%) $\uparrow$ & C-s $\uparrow$ & LPIPS $\downarrow$
& ASR(\%) $\uparrow$ & C-s $\uparrow$ & LPIPS $\downarrow$
& ASR(\%) $\uparrow$ & C-s $\uparrow$ & LPIPS $\downarrow$
& ASR(\%) $\uparrow$ & C-s $\uparrow$ & LPIPS $\downarrow$ \\
\midrule
\multirow{2}{*}{SD-1.5}~\cite{rombach2022high}
& AnyBreak
& 84.51 & 0.26 & 0.0708
& 81.25 & 0.26 & 0.0723
& 2.78  & 0.25 & 0.0892
& 74.53 & 0.25 & 0.0723 \\
& AnyBreak-V
& 53.54 & 0.25 & 0.0723
& 52.84 & 0.25 & 0.0736
& 46.61 & 0.25 & 0.0723
& 49.87 & 0.24 & 0.0736 \\
\midrule
\multirow{2}{*}{SD-2}~\cite{rombach2022high}
& AnyBreak
& 85.39 & 0.26 & 0.0700
& 77.17 & 0.26 & 0.0729
& 6.01  & 0.25 & 0.0713
& 82.17 & 0.26 & 0.0702 \\
& AnyBreak-V
& 52.12 & 0.25 & 0.0724
& 49.72 & 0.25 & 0.0743
& 49.29 & 0.24 & 0.0726
& 48.97 & 0.25 & 0.0720 \\
\midrule
\multirow{2}{*}{P2P}~\cite{brooks2023instructpix2pix}
& AnyBreak
& 61.78 & 0.25 & 0.0170
& 43.04 & 0.25 & 0.0250
& 3.29  & 0.25 & 0.0185
& 57.11 & 0.25 & 0.0170 \\
& AnyBreak-V
& 29.49 & 0.24 & 0.0183
& 25.70 & 0.25 & 0.0257
& 29.19 & 0.24 & 0.0172
& 26.07 & 0.24 & 0.0183 \\
\bottomrule
\end{tabular*}
\label{tab:4_main}
\vspace{-0.4cm}
\end{table*}

\subsection{AnyBreak: A Complete Illustration}
Guided by the foregoing insights, we propose AnyBreak, a universal jailbreak attack framework applicable to any T2I models for any malicious concepts. AnyBreak is built upon two cores: (i) cross-modal malicious concepts transfer and (ii) universal adversarial noise optimization under black-box constraints. The complete workflow is shown in~\cref{fig:3_pipeline}.

\textbf{Malicious Concept Transfer.}
AnyBreak enables cross-modal transfer of malicious concepts and performs attacks with clean textual prompt. Specifically, the attacker employs a generator, such as~\cite{cai2025z}, to edit a malicious concept $c_m$ into a target image $x_t$. Following a strategy similar to~\cref{eq:a_2} (detailed below), we perform semantic alignment between $x_c$ and $x_t$ in pixel space. 
Importantly, pixel-level semantic alignment can steer the latent distribution of the clean image $x_c$ toward the unsafe domain associated with the target image $x_t$. Let $p_c(x)$ denote the distribution of perturbed clean images, and let $\mathcal{M} \subset \mathbb{R}^D$ represent the data manifold induced by malicious images through the reverse SDE process in~\cref{eq:a_1}. Specifically, if the perturbed distribution $p_c$ and the target noise distribution $p_t$ satisfy $\mathrm{KL}(p_c \mid p_t)\le \varkappa$ ($\mathrm{KL}$ denotes Kullback–Leibler divergence), then, as $\delta\to 0$, the probability that generated samples fall within the neighborhood $\mathcal{T}_r(\mathcal{M})$ of the unsafe manifold admits the following lower bound~\cite{sakamoto2024geometry}:
\begin{equation}
\mathbb{P}_{X \sim \rho_0}\big[X \in \mathcal{T}_r(\mathcal{M}_m)\big] \ge (1 - \delta) - \sqrt{\frac{\varkappa}{2}}
\label{eq:c_1}
\end{equation}
In summary, we first transfer the malicious concept $c_m$ from text to a target image $x_t$ via a generator, and then leverage a set of encoders $\{f_i\}_{i=1}^n$ to semantically align $x_c$ with $x_t$ in the pixel space, thereby realizing cross-modal malicious concept transfer.~\cref{eq:c_1} further indicates that reducing $\varkappa$, which corresponds to improving semantic alignment, increases the probability mass of $x_c$ drifting toward unsafe domains during denoising process, consistent with empirical observations. A rigorous proof of~\cref{eq:c_1} can be found in the appendix.

\begin{table*}[t]
\centering
\renewcommand{\arraystretch}{1.15}
\setlength{\tabcolsep}{4pt}
\caption{Performance of different methods on SD-1.5 under various defense mechanisms. Best-performing results are bolded.
}
\begin{tabular*}{\textwidth}{@{\extracolsep{\fill}} c | c c c| c c c| c c c|c c c}
\toprule
\multirow{2}{*}{\textbf{Method}}
& \multicolumn{3}{c|}{\textbf{w / o defense}}
& \multicolumn{3}{c|}{\textbf{Gaussion Noise}}
& \multicolumn{3}{c|}{\textbf{Safety Checker}}
& \multicolumn{3}{c}{\textbf{Negative Prompt}} \\
\cmidrule(lr){2-4}\cmidrule(lr){5-7}\cmidrule(lr){8-10}\cmidrule(lr){11-13}
& ASR(\%) $\uparrow$ & C-s $\uparrow$ & LPIPS $\downarrow$
& ASR(\%) $\uparrow$ & C-s $\uparrow$ & LPIPS $\downarrow$
& ASR(\%) $\uparrow$ & C-s $\uparrow$ & LPIPS $\downarrow$
& ASR(\%) $\uparrow$ & C-s $\uparrow$ & LPIPS $\downarrow$ \\
\midrule
MMA~\cite{yang2024mma}
& 42.21 & 0.24 & 0.3937
& 26.48 & 0.24 & 0.4143
& 39.95 & 0.24 & 0.3894
& 46.30 & 0.23 & 0.3874 \\
MMA-Mask~\cite{yang2024mma}
& 22.21 & 0.24 & 0.3937
& 15.29 & 0.24 & 0.4143
& 19.64 & 0.24 & 0.3894
& 44.77 & 0.23 & 0.3874\\
ADVI2I~\cite{zeng2024advi2i}
& 62.82 & 0.22 & 0.1436
& 50.07 & 0.22 & 0.2421
& 10.27 & 0.22 & 0.1450
& 59.14 & 0.22 & 0.1410 \\
\textbf{AnyBreak}
& \textbf{84.51} & \textbf{0.26} & \textbf{0.0708}
& \textbf{81.25} & \textbf{0.26} & \textbf{0.0723}
& 2.78 & 0.25 & 0.0892
& \textbf{74.53} & \textbf{0.25} & \textbf{0.0723} \\
\textbf{AnyBreak-V}
& 53.54 & 0.25 & 0.0723
& 52.84 & 0.25 & 0.0736
& \textbf{46.61} & \textbf{0.25} & \textbf{0.0723}
& 49.87 & 0.24 & 0.0736 \\
\bottomrule
\end{tabular*}
\label{tab:4_compare}
\vspace{-0.2cm}
\end{table*}

\begin{figure*}[t]
\centering
\begin{minipage}{0.24\textwidth}
  \centering
  \includegraphics[width=\linewidth]{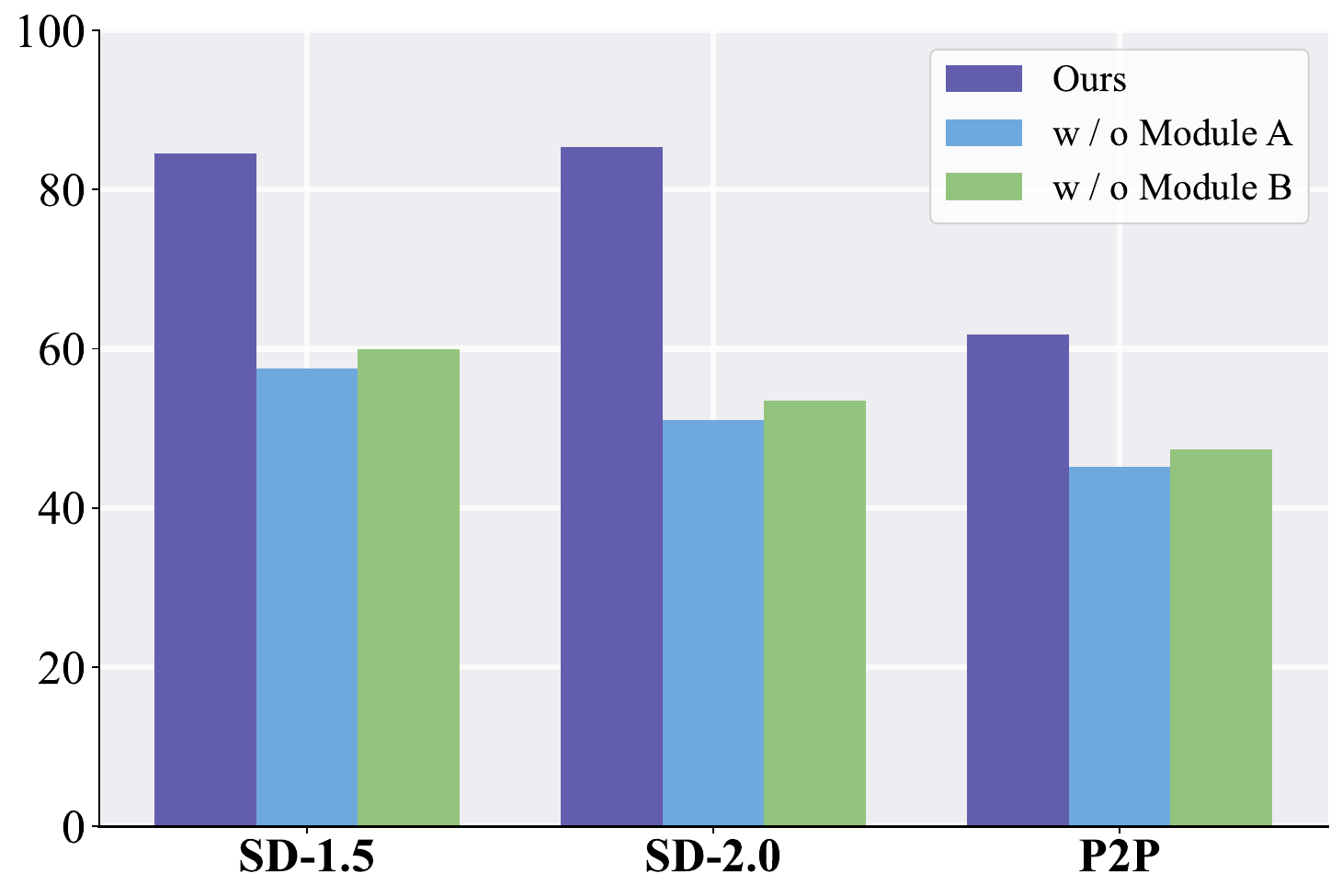}\\[-4pt]
  {\footnotesize (a) module}
\end{minipage}
\hfill
\begin{minipage}{0.24\textwidth}
  \centering
  \includegraphics[width=\linewidth]{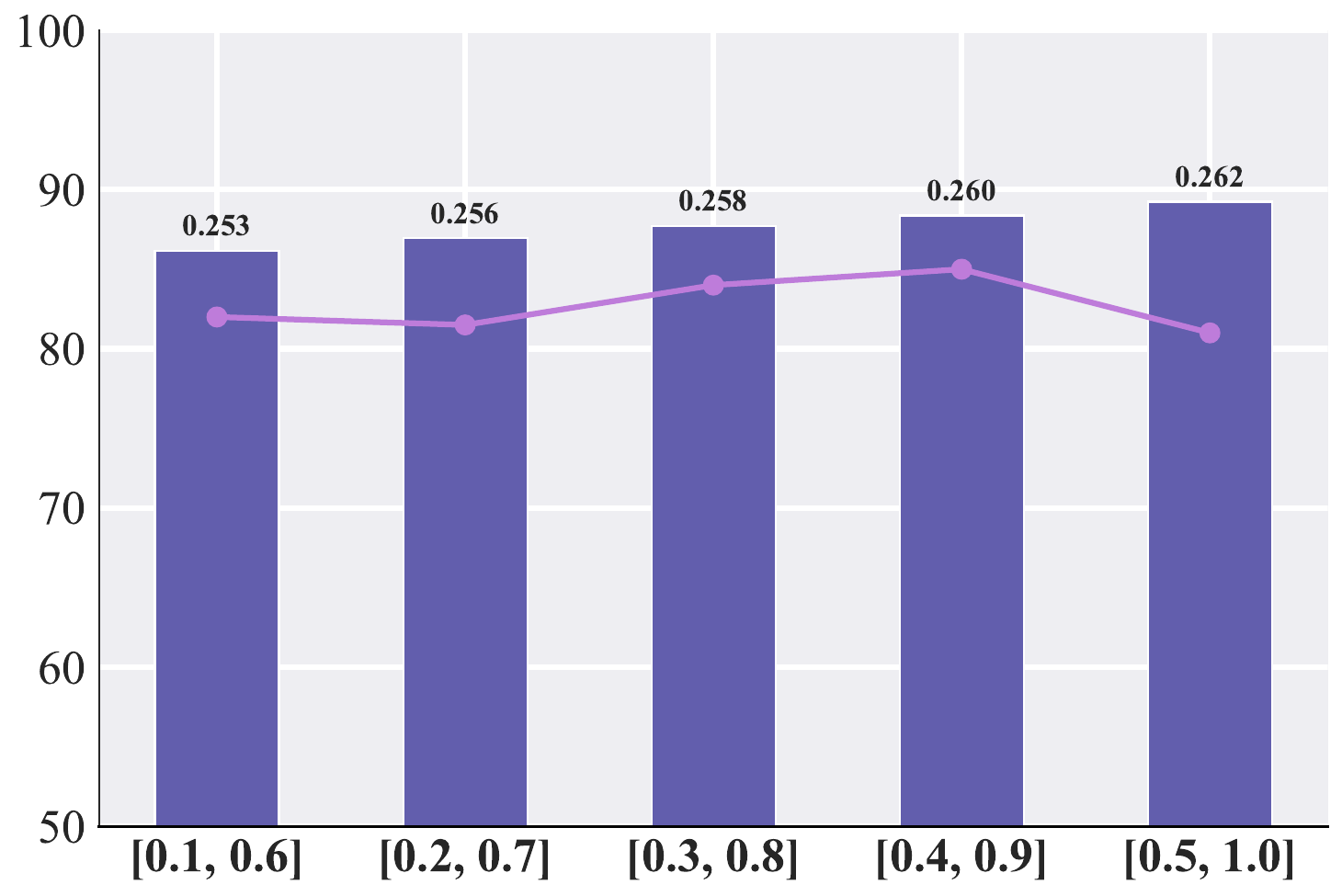}\\[-4pt]
  {\footnotesize (b) window ratio $r$}
\end{minipage}
\hfill
\begin{minipage}{0.24\textwidth}
  \centering
  \includegraphics[width=\linewidth]{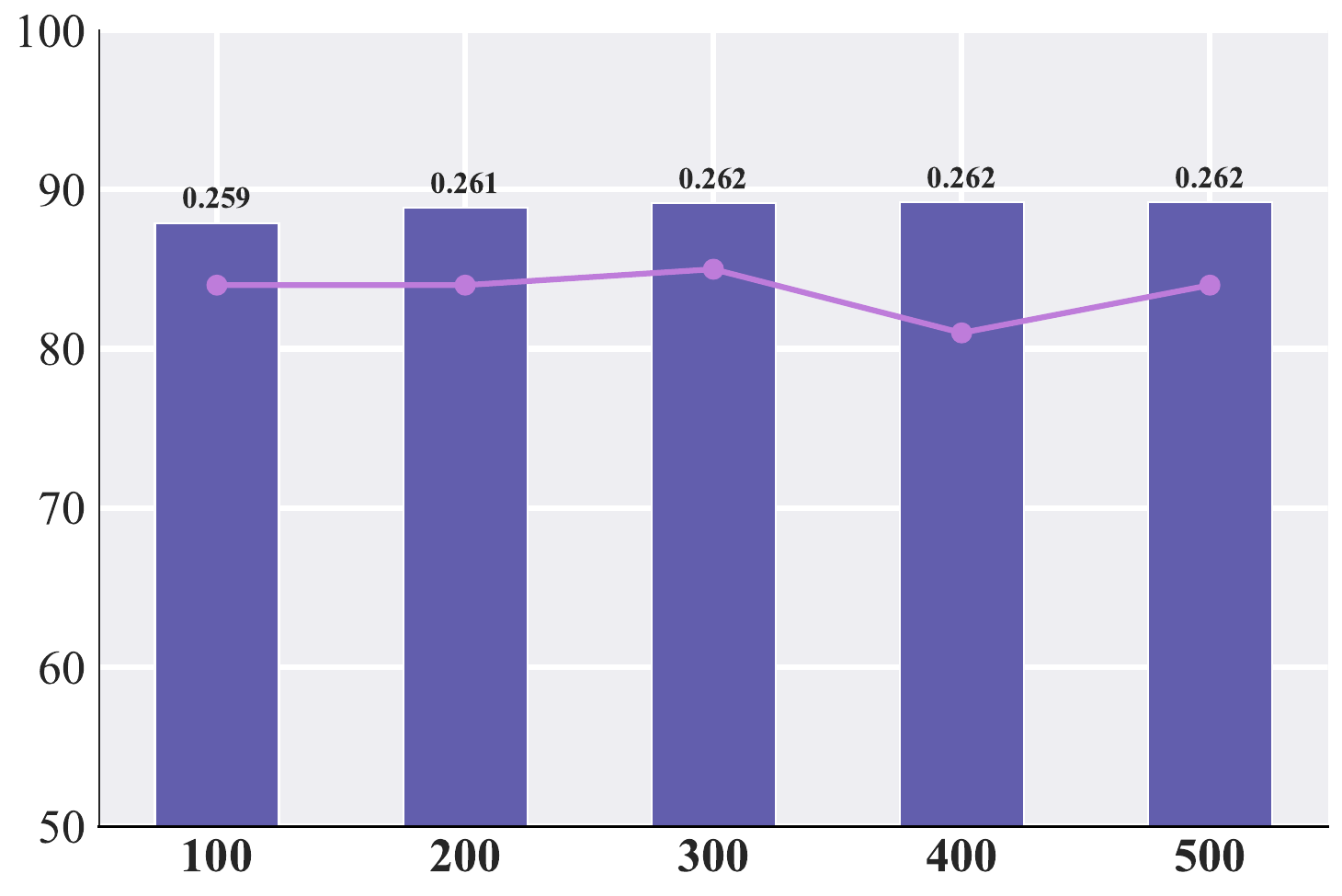}\\[-4pt]
  {\footnotesize (c) optimization iteration $K$}
\end{minipage}
\hfill
\begin{minipage}{0.24\textwidth}
  \centering
  \includegraphics[width=\linewidth]{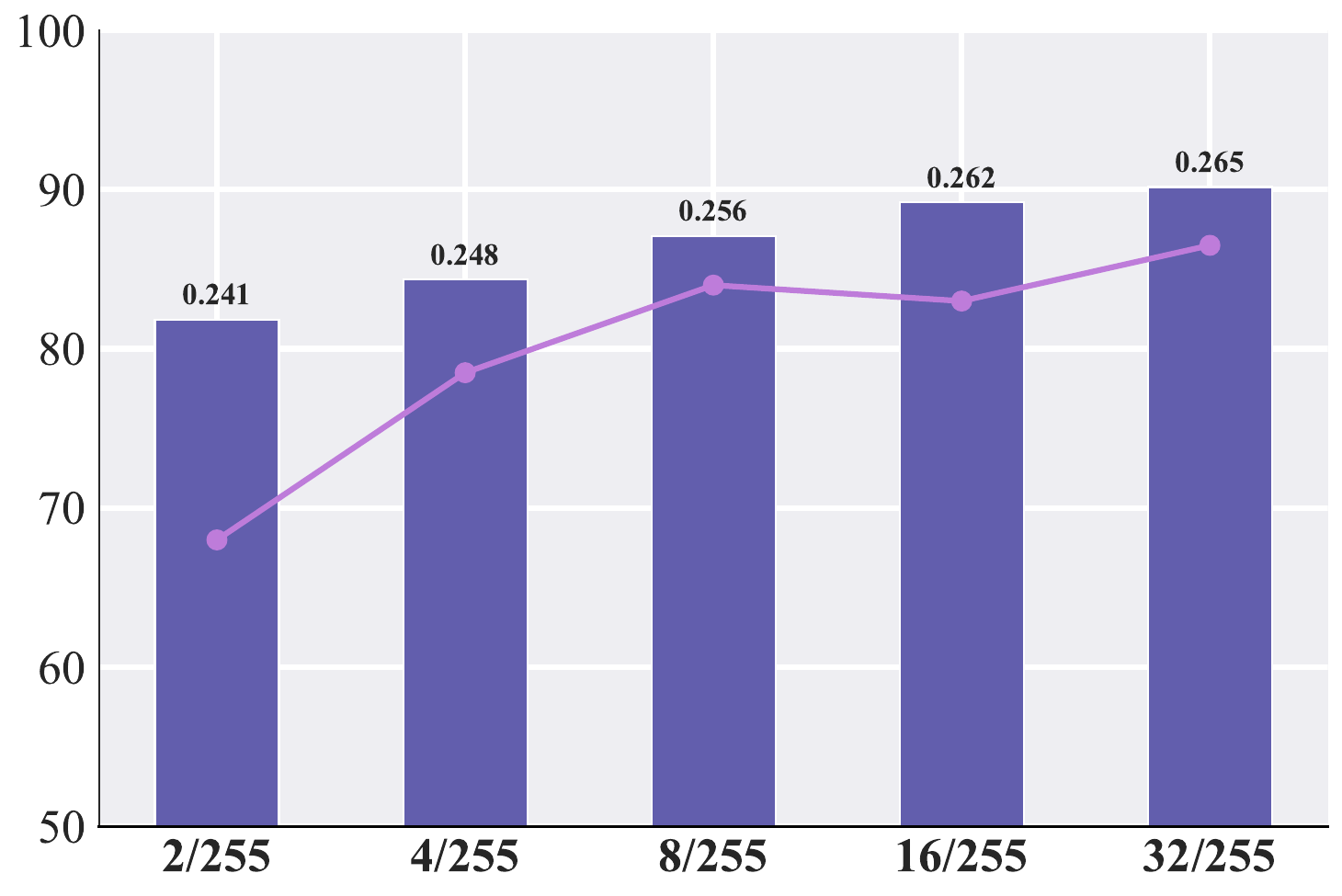}\\[-4pt]
  {\footnotesize (d) perturbation budget $\Delta$}
\end{minipage}
\caption{Ablation study on hyperparameters. In figure (b)–(d), bars denote CLIP-score, while lines indicate ASR.}
\label{fig:4_ablation}
\vspace{-0.4cm}
\end{figure*}

\textbf{Universal Adversarial Noise Optimization.}
As discussed in Insight II, optimizing global similarity alone overlooks local structural discrepancies, which can induce the collapse of adversarial semantic. To address this limitation, we introduce a random local-window alignment mechanism that conducts $K$ rounds of stochastic feature alignment over spatial subregions, enabling multi-scale and progressive optimization while maintaining global coherence. Specifically, in each round, we construct a set of sliding-window candidates over the spatial dimensions. Given a window size $(h, w)$ and stride $v$, all valid windows are indexed by their top-left coordinates:
\begin{equation}
\mathcal{G} = \{(i,j) \in [0,H\!-h] \times [0,W\!-w] \mid i,j \equiv 0\!\!\!\! \pmod{v}\}.
\label{eq:c_2}
\end{equation}
Since the encoder operates on square inputs, we set $h = w$ in practice and define the cropping ratio as $r = \frac{h}{H}$. At each optimization iteration, we independently sample local windows from $\mathcal{G}$ for the clean image $x_c$ and the target image $x_t$, and apply the corresponding crops separately. The cropping operators are denoted by $\mathcal{C}{r_c}$ and $\mathcal{C}{r_t}$, where $r_c$ and $r_t$ represent the cropping ratios for the clean and target images, respectively, with $0< r_c, r_t < 1$.
We optimize the adversarial noise using \textit{fast gradient sign method} (FGSM)~\cite{goodfellow2014explaining} on a set of image encoders $\{f_i\}_{i=1}^n$ as:
\begin{equation}
\max_{\delta}\ \frac{1}{n}\sum_{i=1}^n\mathrm{Cos}\Big(f_i\big(\mathcal{C}_{r_c}(x_c+\delta)\big),\ f_i\big(\mathcal{C}_{r_t}(x_t)\big) \Big).
\label{eq:c_3}
\end{equation}
Here, $\mathrm{Cos}$ denotes cosine similarity, and $|\delta|_\infty \le \Delta$. Through progressive iterations, the adversarial perturbation gradually captures both global semantics and local discrepancies.
\subsection{AnyBreak-V: Evasion on safety checker}
In practical deployments, T2I models are typically equipped with a post-hoc safety checker (\cite{compvis2022safetychecker, rando2022red}) after the generation stage. This checker compares the generated image against a set of predefined NSFW concept embeddings using a similarity metric, and blocks the output once the similarity exceeds a prescribed threshold. Our empirical results show that this mechanism can be effectively circumvented by introducing an additional compensation term into AnyBreak, yielding its variant \textit{\textbf{AnyBreak-V}}, which successfully evades the safety checker. Specifically, we simulate the checker’s operating principle by following SLD~\cite{schramowski2023safe} to select $M$ images that contain malicious concepts, such as pornographic or violent content, and extract their embeddings $\{c_i\}_{i=1}^M$. Based on~\cref{eq:c_3}, these embeddings are incorporated as a compensation term that explicitly drives the generated image away from the malicious concept embedding space:
\begin{equation}
\max_{\delta}\ -\frac{1}{M}\sum_{i=1}^M\mathrm{Cos}\Big(f_i(x_c+\delta),\ c_i\Big).
\label{eq:c_4}
\end{equation}
During the adversarial noise optimization stage, AnyBreak-V is obtained by directly incorporating~\cref{eq:c_4} into the objective in~\cref{eq:c_3} on top of AnyBreak.
\section{Experiments}
\label{sec:experiments}


\begin{table}[t]
\centering
\caption{ASR(\%) across Different Malicious Concepts.}
\label{tab:malicious_concept}
\setlength{\tabcolsep}{9pt}
\begin{tabular}{c c c c c c}
\toprule
\textbf{Model} & \textbf{Adult} & \textbf{Bloody} & \textbf{Violence} & \textbf{Horror} & \textbf{Racism} \\
\midrule
\addlinespace[2pt]
SD-1.5 & 84.5 & 83.5 & 82.5 & 71.0 & 87.5 \\
SD-2.0 & 86.5 & 79.5 & 80.5 & 66.0 & 78.5 \\
\bottomrule
\end{tabular}
\vspace{-0.4cm}
\label{tab:4_concept}
\end{table}

\subsection{Experimental Settings}
\noindent{\textbf{Datasets.}} We conduct experiments on the "sexy" category of the NSFW Data Scraper~\cite{nsfw_data_scraper}.

\noindent{\textbf{Models.}} 
We evaluate on SD-V1.5-Inpainting (SD-1.5), SD-V2-Inpainting (SD-2) and InstructPix2Pix (P2P).

\noindent{\textbf{Defenses.}}
We evaluate the robustness of AnyBreak under multiple defense mechanisms. As an input-level defense, we inject random Gaussian noise into adversarial images to disrupt optimized perturbations~\cite{honig2024adversarial}. As a passive output-side defense, we employ a post-hoc safety checker~\cite{compvis2022safetychecker}. We further include Negative Prompt~\cite{rombach2022high} as a concept erasure mechanism applied during the denoising process to erase malicious concepts.

\noindent{\textbf{Metrics.}}
We report \textit{attack success rate} (ASR) to measure attack effectiveness, defined as the number of $x_{\mathrm{nsfw}}$ that successfully jailbreak into NSFW outputs. We also evaluate image quality. Specifically, \textit{CLIP-score} (abbr. C-s)~\cite{radford2021learning} measures the consistency between $x_{\mathrm{nsfw}}$ and the original malicious textual concept $c_m$, while LPIPS~\cite{zhang2018unreasonable} assesses perceptual quality.

\noindent{\textbf{Parameter Setting.}}
In our experiments, we set $r_c,r_t \in [0.5, 1.0]$, the iterations $K = 300$, the perturbation budget $\Delta = 16/255$, the standard deviation of the Gaussian noise to $0.05$, and employ ViT-B/32 for CLIP-score computation.

\begin{figure}[!t]
    \centering
    \includegraphics[scale=0.54]{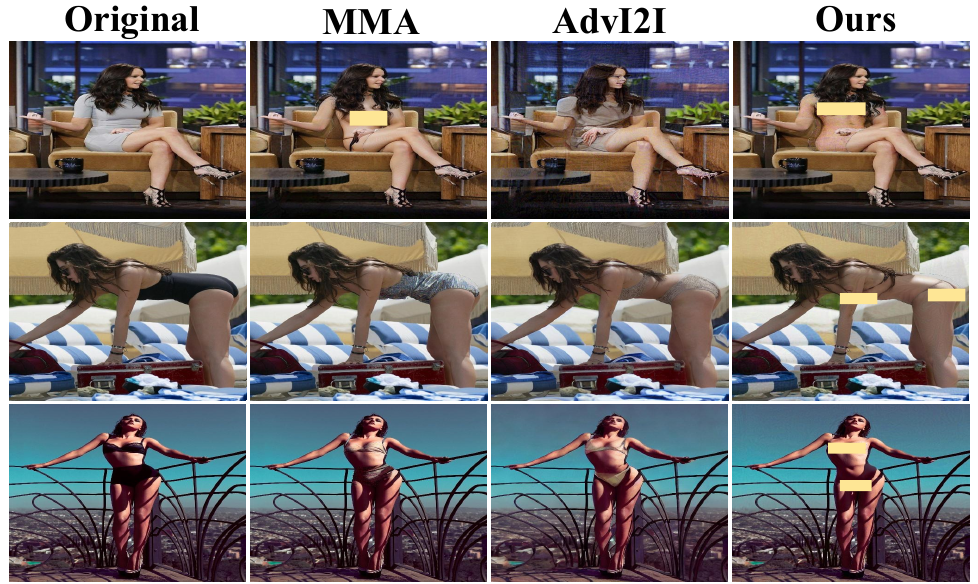}
    \vspace{-0.4cm}
    \caption{Visualization of jailbreak output images across different methods.
    }
    \label{fig:4_visual}
    \vspace{-0.6cm}
\end{figure}

\subsection{Results and Analysis}
As shown in~\cref{tab:4_main}, AnyBreak demonstrates consistently strong performance across multiple models. Without defenses, it achieves an average black-box ASR of up to 77\%. AnyBreak-V underperforms AnyBreak, as the compensation loss steers perturbation away from certain malicious concepts, thereby reducing adversarial semantic effectiveness, at the cost of enhanced ability to bypass safety detectors. AnyBreak exhibits strong robustness, with the average ASR decreasing by less than 10\% under random Gaussian noise and negative prompt. While safety checkers provide effective protection, AnyBreak-V evades these defenses via the compensation loss, resulting in an average ASR drop of less than 5\% relative to the no-defense setting. Moreover, AnyBreak achieves an average C-s of up to 0.25, compared to approximately 0.28 measured directly between input prompts and generated images of T2I models. The average LPIPS is as low as 0.05, indicating that $x_{\mathrm{nsfw}}$ preserves high visual fidelity. We  also evaluate generalization across malicious concepts, as shown in~\cref{tab:4_concept}, AnyBreak enables universal attacks across concepts. Experimental details can be found in the appendix.

\subsection{Comparison Study}
We select ADVI2I~\cite{zeng2024advi2i} and MMA~\cite{yang2024mma} as representative image-side jailbreak baselines. For fair comparison, we remove text-level sensitive keywords from MMA, referred to
as MMA-mask. As shown in~\cref{tab:4_compare}, AnyBreak consistently outperforms all baselines across multiple defense strategies in both ASR and image generation quality. Notably, under safety checker deployment, where all baselines fail, AnyBreak-V still achieves a high ASR. Visual results in~\cref{fig:4_visual} further confirm that AnyBreak induces NSFW content more effectively while maintaining superior visual fidelity.

\subsection{Ablation Study}
\noindent{\textbf{Impact of module.}}
We conduct ablation studies on concept transfer (Module A) and multi-scale noise optimization (Module B), as shown in~\cref{fig:4_ablation}(a). When module A is removed, the ASR decreases substantially across all models. Without module B, consistent with Observation II, similarly results in a notable ASR degradation.

\noindent{\textbf{Impact of $r$.}}
~\cref{fig:4_ablation}(b) evaluates different ranges of $r$ while keeping the overall variation span fixed. The results indicate that the multi-scale noise strategy is highly robust and exhibits low sensitivity to the choice of window size.

\noindent{\textbf{Impact of $K$.}}
As shown in~\cref{fig:4_ablation}(c), adversarial optimization converges within a relatively small number of iterations, and increasing $K$ has limited influence on the final performance.

\noindent{\textbf{Impact of $\delta$.}}
~\cref{fig:4_ablation}(d) shows that ASR increases with a larger perturbation budget. However, excessive perturbations degrade image generation quality, as reflected by a decrease in LPIPS.
\section{Conclusion}
\label{sec:Conclusion}
In this paper, we propose AnyBreak, a jailbreak framework that targets any T2I models for any concepts for NSFW generation. AnyBreak enables cross-modal transfer of malicious concepts by constructing universal adversarial noise through multi-scale semantic alignment, allowing successful jailbreak attacks using only cleanprompts and adversarial images at inference time. We further introduce the variant AnyBreak-V, designed to evade safety checkers. Our findings expose a critical security vulnerability in current T2I models: even with deployed prompt filters, models can still be induced to generate NSFW content through adversarial images.

\bibliographystyle{IEEEbib}
\bibliography{icme2026references}
\end{document}